\documentclass[letterpaper,twocolumn,10pt]{article}
\usepackage{usenix-2020-09}
\usepackage{booktabs}
\usepackage{amsmath,amssymb}
\usepackage{times}  
\usepackage{tikz}
\usepackage{url}
\usetikzlibrary{positioning,arrows.meta,shapes.geometric,calc}

\usepackage[T1]{fontenc}
\usepackage{cmap}
\usepackage[utf8]{inputenc}

\begin{document}

\title{
The Abstention Protocol: RCA for Clos Fabrics (Operational Systems)}

\author{
{\rm Madhava Gaikwad}\thanks{This work was performed when Madhava Gaikwad was
with Microsoft.}\\
Independent
\and
{\rm Deepak Pandey}\\
Microsoft
}

\maketitle
\begin{abstract}
Root cause analysis (RCA) in large datacenter networks is challenging because
telemetry is noisy, partial, and asynchronous. Score-based approaches degrade
under these conditions, often yielding unstable or incorrect attributions.

We present \textsc{CoreSec}, a production RCA system that replaces weighted
fusion with a PAM-style abstention algebra. Telemetry agents are composed using
control flags that yield deterministic decisions and explicit abstention when
evidence is ambiguous. CoreSec combines this algebra with topology-aware
configurations that capture failure surfaces across Clos fabrics and converge
monotonically as evidence accumulates.

Deployed at hyperscale, CoreSec provides stable and explainable RCA behavior
across diverse environments without retuning. Our experience shows that
structured composition with abstention forms a practical foundation for
automated RCA in real-world cloud networks.
\end{abstract}
\section{Introduction}

\begin{figure}[t]
  \centering
  \includegraphics[width=0.9\columnwidth]{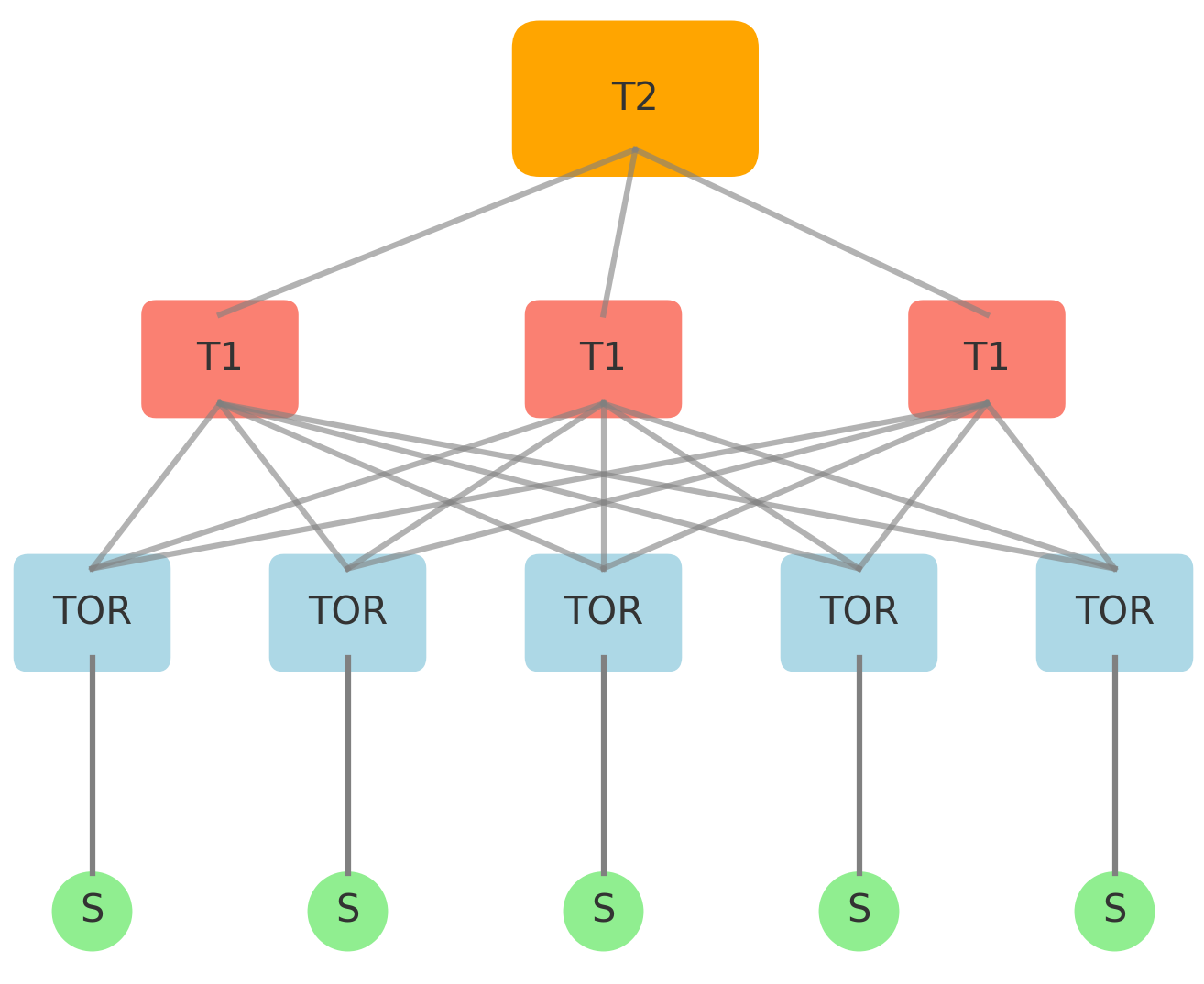}
  \caption{Three-tier Clos topology. Servers connect to TORs, TORs to T1
  aggregation switches, and T1s to T2 spines.}
  \label{fig:clos}
\end{figure}

Large cloud networks operate with continuous background faults. In a
cluster with thousands of switches, it is normal to see a server-to-TOR
(top-of-rack switch) cable with CRC errors, a switch-to-switch link that
flaps intermittently, or a TOR in the middle of an upgrade. (A link
\emph{flaps} when it rapidly transitions between up and down states.)
Clos topologies are multi-stage switch fabrics that provide many
equal-cost paths between any two endpoints (Figure~\ref{fig:clos}).
They absorb these faults through path diversity, so the fabric continues
forwarding traffic. Prior studies of hyperscale networks report the
same behavior~\cite{guo2015pingmesh}. But this background noise makes
root cause analysis difficult: when a customer workload fails, many
entities show faults. Only a subset are related to the incident.
Most are routine background failures.

Accurate RCA matters for two reasons. Customers expect precise explanations
for connectivity disruptions: what failed, whether it will recur, and what
Azure is doing about it. Engineering teams use aggregated RCA data to identify
systemic weaknesses in firmware, optics, and operational processes. Patterns
across thousands of incidents reveal which hardware families fail most often
and which operational changes introduce
risk~\cite{googleSREbook,awsPostmortems,gcpIncidents}.

Most failures CoreSec attributes are not fail-stop. They are
\emph{gray failures}~\cite{huang2017gray}: partial, probabilistic, or
intermittent malfunctions. A device with a loose optical module may drop
two percent of packets on one direction of one link. A linecard with a
memory bit-flip may corrupt headers only for flows that hash to a
particular ECMP path. A firmware bug may periodically reboot a switch
and recover before health monitors notice. Their surface symptoms blend
into the steady background fault rate (\S\ref{sec:background}), and a
fresh on-call engineer cannot tell, just from looking at counters and
probes, whether a particular signal indicates the incident under
investigation or one of the dozens of unrelated faults always quietly
happening somewhere in the fabric.

CoreSec runs on top of an existing operational pipeline at Azure. Many
of its inputs come from telemetry systems built and deployed before
CoreSec, two of which are particularly relevant.
Pingmesh~\cite{guo2015pingmesh} runs on every server and continuously
measures end-to-end latency and packet loss between server pairs across
the fabric. It tells operators \emph{that} something is wrong and
\emph{which region} is affected, without saying which device is at
fault. NetBouncer~\cite{tan2019netbouncer} sits a layer below: it
actively probes paths through the Clos fabric and infers \emph{which
links and devices} are unhealthy. Neither answers the question we kept
getting paged for: given a specific incident raised against a specific
customer service, which network entity \emph{caused it}? At any moment
many entities will be unhealthy, most unrelated to the incident under
investigation. CoreSec consumes the outputs of Pingmesh, NetBouncer,
and a number of other agents (device counters, traffic-derived signals,
control-plane events, infrastructure health summaries) and resolves
them into a single attribution. We discuss the relationship to these
systems in more detail in \S\ref{sec:related}.

Earlier RCA systems at Azure used weighted aggregation of telemetry signals.
A \emph{telemetry agent} is a software component that observes some
property of the network (active probes, device counters, or traffic
flows) and reports a per-entity verdict on whether that property looks
healthy. Each agent produced a score, and the entity with the highest
weighted sum was blamed. This approach failed in predictable ways. Background faults always
contributed some signal, so the system found a culprit even when the network
was not at fault. Tuning weights to reduce false positives for one failure
mode increased them for another. The false positive rate fluctuated between
18 and 22 percent and could not be
controlled~\cite{tan2019netbouncer,kanuparthy2016ytrace,guo2015pingmesh}.

The key observation behind CoreSec is that RCA in this regime is a
composition problem. No single telemetry agent is reliable across all
failure modes: active probes detect link failures quickly but miss
software defects; device counters catch hardware degradation but
produce noise; traffic-derived signals reflect customer impact but
have sparse coverage. We describe these agent classes in detail in
\S\ref{sec:background} (\S2.2). CoreSec assigns each agent a control
flag that specifies whether its evidence is required, sufficient, or
optional. When evidence conflicts or is missing, the system abstains.
This structure is inspired by Pluggable Authentication Modules
(PAM)~\cite{samar1996unified}, which combines independent
authentication checks (password, biometrics, hardware token) by tagging
each one as required, sufficient, or optional. The composition problem
is the same one we face with telemetry.

This paper makes four contributions:

\begin{itemize}
    \item \textbf{A composable RCA algebra with explicit abstention.}
    We adapt the control-flag model from PAM to network RCA. Each
    telemetry agent is assigned a flag: requisite, required, sufficient,
    or optional. Five configurations run in parallel, each targeting a
    different failure surface in the Clos topology. The algebra admits
    a third outcome alongside healthy and unhealthy --- the system
    abstains when evidence is missing or conflicting --- and produces
    deterministic, explainable decisions.

    \item \textbf{Topology-aware heuristics.} We derive a small set of
    topology thresholds for server-to-TOR, TOR-to-T1, and cluster-level
    attribution from multi-year analysis of historical
    incidents~\cite{zhu2017conga,roy2015inside}. These thresholds (described
    in \S\ref{sec:hierarchy}) sit at the knees of their error curves and
    generalize across various Azure deployments without recalibration.

    \item \textbf{Three-year production deployment.} CoreSec has processed
    over 700,000 incidents and reduced false positives from 18-22 percent
    to below 1 percent. We treat abstentions as explicit false negatives;
    this rate dropped from 10 percent initially to 1.5 percent over the
    most recent six months. The system eliminated three full-time engineers
    worth of manual RCA work.

    \item \textbf{Formal model.} We give an algebraic specification of
    the merge operator as a three-valued lattice with a short
    associativity proof, together with absorption and consensus
    properties (Appendix~\ref{app:algebra}) that account for CoreSec's
    determinism and order-invariance under asynchronous telemetry.
\end{itemize}

\begin{figure}[t]
\centering
\begin{tikzpicture}[
  font=\footnotesize,
  every node/.style={inner sep=2pt},
  box/.style={draw, rounded corners=1pt, align=center,
              minimum width=12mm, minimum height=6mm},
  cfg/.style={box, fill=gray!8, minimum width=14mm, minimum height=8mm},
  decide/.style={box, fill=gray!15, minimum width=32mm},
  result/.style={box, fill=gray!20, minimum width=18mm},
  arrow/.style={-{Stealth[length=4pt]}, thin},
]
\node[box] (incident) at (0, 3.6) {Incident alert};

\node[cfg] (c1) at (-3.2, 2.0) {Config 1\\\scriptsize svr--TOR};
\node[cfg] (c2) at (-1.6, 2.0) {Config 2\\\scriptsize TOR};
\node[cfg] (c3) at ( 0.0, 2.0) {Config 3\\\scriptsize sw--sw};
\node[cfg] (c4) at ( 1.6, 2.0) {Config 4\\\scriptsize T1};
\node[cfg] (c5) at ( 3.2, 2.0) {Config 5\\\scriptsize T2};

\node[decide] (merge) at (0, 0.4) {Hierarchy heuristics};

\node[result] (att) at (-1.4, -1.1) {Attribution};
\node[result] (abs) at ( 1.4, -1.1) {Abstain};

\foreach \c in {c1,c2,c3,c4,c5} {
  \draw[arrow] (incident.south) -- (\c.north);
  \draw[arrow] (\c.south) -- (merge.north);
}

\draw[arrow] (merge.south) -- (att.north);
\draw[arrow] (merge.south) -- (abs.north);

\node[align=left, font=\scriptsize\itshape, anchor=west]
  at (-3.6, 0.4) {re-run\\every 5 min\\for 16 min};
\end{tikzpicture}
\caption{CoreSec dataflow. An incident triggers five parallel
configurations; hierarchy heuristics combine their votes into one
attribution or an abstention.}
\label{fig:pipeline}
\end{figure}
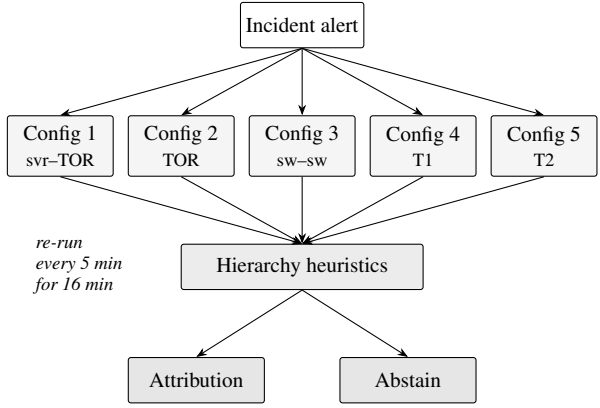

CoreSec is invoked post-incident, not as a continuous monitor: an
external detector raises an alert, and CoreSec attributes that specific
incident from a sixteen-minute telemetry window. Each of its five
configurations targets one component type in the Clos topology
(server-to-TOR cable, TOR, switch-to-switch cable, T1, T2). Within a
configuration the PAM-style algebra resolves agent verdicts to
\emph{healthy}, \emph{unhealthy}, or \emph{indeterminate}; across
configurations, topology heuristics (\S5.2) decide which layer is
responsible when several vote. If no configuration has sufficient
evidence, CoreSec abstains. Figure~\ref{fig:pipeline} shows this flow.

\textbf{A walkthrough.} To make the flow concrete: suppose three servers
fail in a way that points to upstream trouble. Configuration~2 (TOR
switch) marks each server's TOR as a candidate. Configuration~4 (T1
switch) checks whether enough TORs in the same T1's fan-out are also
unhealthy. If at least two-thirds are, the T1 votes and suppresses the
individual TOR candidates. The final RCA is the T1. \S\ref{sec:hierarchy}
gives the full numbers.
\section{Background and Motivation}
\label{sec:background}
\textit{Clos Fabrics in Hyperscale Networks}
Hyperscale cloud networks use multi-stage Clos topologies because they offer
predictable bandwidth, uniform latency, and clear fault containment. A three
tier fabric connects servers to top-of-rack (TOR) switches, TORs to aggregation
switches (T1), and T1s to spine switches (T2). Each server typically connects
to one or more TORs for fault tolerance, and each TOR has several uplinks to
independent T1s. This structure creates many equal-cost paths and allows traffic
to be rerouted quickly when a link or device fails~\cite{guo2015pingmesh,roy2015inside}.

Clos fabrics experience a continuous background rate of faults. Optical modules
drift, cables accumulate CRC errors, control plane processes restart, and
firmware upgrades proceed on rolling schedules. Studies of deployed clusters
show hundreds of transient link faults and dozens of device reboots daily in a
single region~\cite{roy2015inside}. Multipath routing absorbs these faults, so
they rarely cause visible disruption. It is normal for 0.3 to 1 percent of
links to show loss, flaps, or degraded optics at any given
time~\cite{guo2015pingmesh}.

This baseline complicates root cause analysis. A customer-visible incident
often coincides with several unrelated background faults. The topology shapes
how failures propagate: a TOR failure typically affects only its rack, while a
T1 failure creates correlated symptoms across many TORs. CoreSec's heuristics
are built around these propagation patterns.

\textbf{Telemetry Agents and Failure Signals}

Large-scale Clos networks rely on multiple telemetry sources to detect and
localize failures. Each source observes the network from a different vantage
point and reports at different temporal and spatial
resolutions~\cite{tan2019netbouncer,guo2015pingmesh,kanuparthy2016ytrace}.

\textit{Active probes.}
Systems like Pingmesh~\cite{guo2015pingmesh} inject synthetic probes along
controlled paths and measure loss and latency. Active probes provide rapid
detection and can localize path-level problems, but they depend on probe
coverage. Some links may see sparse or no probing during certain intervals.

\textit{Device counters.}
Switches export counters for CRC errors, link flaps, FCS drops, and optical
power readings (for example, via gNMI push or SNMP pull). These provide direct evidence of hardware issues but update
at irregular intervals and are sensitive to transient
noise~\cite{roy2015inside}. Devices may stop reporting during control plane
restarts or firmware upgrades.

\textit{Traffic-derived signals.}
Systems such as NetBouncer~\cite{tan2019netbouncer} analyze paths taken by
real traffic to detect links whose behavior deviates from baseline. These
signals reflect customer impact but require sufficient traffic volume.

\textit{Infrastructure signals.}
Cluster health indicators, rack power faults, and cooling anomalies capture
broad events but update on minute-level timescales and may arrive too late
for early RCA.

No single agent is reliable across all failure modes. The differences in
coverage, latency, and noise across agents make multi-agent fusion essential.

\textit{Operational Requirements}

Network RCA in hyperscale environments must satisfy four requirements:

\begin{itemize}
    \item \textbf{Controllability.} Operators must be able to bound the false
          positive rate.
    \item \textbf{Explainability.} Each decision must be explainable in terms
          of agent outputs.
    \item \textbf{Extensibility.} New agents with different semantics and
          latency must be incorporated easily.
    \item \textbf{Topology awareness.} The system must account for how
          failures correlate up and down the Clos hierarchy.
\end{itemize}

Prior fault-localization systems address some of these requirements but not
all in a single design~\cite{tan2019netbouncer,kanuparthy2016ytrace,arzani2018007}.

\textit{From Incidents to Design}

CoreSec emerged from a multi-year review of incorrect RCAs. Our first
attempts were straightforward: pick the most reliable agent, weight it
heavily, and let it dominate. Within months we saw why that does not
work. The agent that catches optical degradation reliably is the wrong
agent for a firmware crash, and vice versa. Reliability turned out to
be failure-specific, not agent-specific.

We also had to drop several infrastructure signals from the decision
path because they introduced feedback loops with CoreSec itself
(\S\ref{sec:hierarchy} gives the details). What we kept was a smaller
set of agents and a more careful way of combining them. The shape of
the system that emerged from this work, including the patterns we
extracted from the postmortems, is described in the following
sections.
\section{PAM-Style Composable RCA Algebra}

CoreSec builds its root cause analysis logic on a composable algebra inspired by
Pluggable Authentication Modules (PAM) \cite{samar1996unified,geisshirt2007pluggable}.
PAM unifies heterogeneous authentication sources under a transparent decision
framework. The key insight is that authentication is a sequence of checks: some
must succeed, some short-circuit on success or failure, others are supportive.
For example, in a typical PAM stack a successful biometric check can
short-circuit the rest of the stack to grant access; a failed mandatory
password check denies access regardless of what the rest of the stack reports;
an optional usage-history check only matters when the mandatory checks have
not produced a verdict. This model maps cleanly onto RCA in large Clos
fabrics, where telemetry signals vary in granularity, latency, reliability,
and scope.

Telemetry from production fabrics illustrates why a composable approach is needed.
Faults are frequent but rarely catastrophic \cite{meza2018large, tan2019netbouncer}:
a steady background of link flaps, device reboots, optical degradation,
and partial loss, masked by Clos redundancy. When noise accumulates, score-based
RCA destabilizes. Active probing and passive signals face limited coverage,
missing data, and inconsistent correlation between symptoms and root cause
\cite{herodotou2014scalable, xin2022data}. Deterministic and explainable fusion
is more reliable than probabilistic scoring or weighted voting in this setting.

CoreSec’s algebra provides this logic. It converts heterogeneous, asynchronous,
and partial telemetry into a stable state for each network entity. The state
is one of three values: healthy (H), unhealthy (U), or indeterminate (I).
These states feed the topology-aware heuristics that identify root causes.

\textit{Control Flags}

Each telemetry agent is assigned one control flag. The flag defines how the
agent’s data influences RCA:

\begin{itemize}
  \item \textbf{Requisite}: If this agent returns stale data or fails its threshold,
        the configuration abstains immediately. This prevents decisions when
        essential evidence is missing.
  \item \textbf{Required}: All required agents must pass for the configuration to vote.
        If any required agent fails, the configuration abstains at the end.
  \item \textbf{Optional}: This agent contributes only when requisite and required
        agents do not produce a decisive result. Its evidence supports but does not
        determine a vote.
  \item \textbf{Sufficient}: A pass from this agent triggers an immediate vote. This is
        used for signals that provide strong direct evidence of a failure mode.
\end{itemize}

These flags capture empirical differences among agents. Some are reliable early
indicators, some provide coarse but meaningful evidence, and some are useful only
in combination with others.

To make this concrete, consider how the server-to-TOR cable configuration
(\S\ref{sec:configs}) assigns flags. A loss-of-carrier signal from the
server NIC is marked \emph{requisite}: if this signal is missing or stale,
the configuration abstains. An RDMA timeout (an
end-to-host signal that the application could not reach the network) is
marked \emph{sufficient}: its presence alone is decisive evidence of
connectivity loss between the server and its TOR. CRC counter trends are
marked \emph{required}: they must be consistent with a fault but cannot
trigger a vote on their own. Recent control-plane events on the TOR are
marked \emph{optional}: they support but do not drive the decision. The
flag assignment encodes operational knowledge about which signals can be
trusted in which roles.

\textit{Evaluation Logic and Freshness}

Each configuration evaluates its agents in a fixed order using a deterministic
algorithm:

\begin{verbatim}
required_ok = True
for agent in config.agents:
    v = agent.verdict()  # pass | fail | abstain
    if agent.flag == SUFFICIENT and v == pass:
        return VOTE
    if agent.flag == REQUISITE and v == fail:
        return ABSTAIN
    if agent.flag == REQUIRED and v != pass:
        required_ok = False
return VOTE if required_ok else ABSTAIN
\end{verbatim}

Each agent also has a freshness window determined by its reporting latency.
Data older than the window is ignored. Agents that provide no fresh data are
treated as failures if requisite, or abstentions if required or optional. This
addresses a common issue in telemetry systems where staleness and missing reports
occur during partial outages or control plane churn \cite{tan2019netbouncer,
xin2022data}.

\textit{Per-Agent Thresholds and Abstention Region}

Telemetry metrics vary widely in noise. Counter spikes can come from traffic
bursts and not from real failures. Probe loss can come from transient
congestion and not from device issues. Each agent $r$ has a raw signal
$s_r(e)$ for entity $e$ and two thresholds, $\theta_r^-$ (below which the
signal looks healthy) and $\theta_r^+$ (above which the signal looks
faulty):

\[
\text{result}_r(e) =
\begin{cases}
\textsf{pass}, & s_r(e) < \theta^-_r, \\
\textsf{fail}, & s_r(e) \ge \theta^+_r, \\
\textsf{abstain}, & \text{otherwise}.
\end{cases}
\]

The abstain region absorbs benign fluctuations. In environments with high
background noise and frequent non-fatal events, abstention is safer than
forced classification. Large-scale reliability studies report the same
conclusion \cite{meza2018large}.

\textit{Why PAM-Style Composition Provides Operational Strength}

The algebra offers three advantages in operational cloud networks:

\begin{itemize}
  \item \textbf{Controllability}. Requisite flags gate decisions and thresholds bound
        sensitivity. Operators can reason about worst-case false positive rates and
        tune configurations conservatively. This is difficult with weighted or
        probabilistic RCA, where error bounds degrade under missing data or
        distribution shift \cite{harsh2023flock}.
  \item \textbf{Explainability}. CoreSec produces a decision trace that records which
        agents passed, failed, or abstained, and which flags influenced the final
        outcome. These traces support auditing and postmortem analysis.
  \item \textbf{Extensibility}. New telemetry agents can be integrated by assigning a
        control flag. Existing configurations do not need to change. This simplifies
        maintenance as instrumentation evolves.
\end{itemize}

Additional algebraic details are provided in Appendix~\ref{app:algebra}.
\section{Composable RCA Configurations}
\label{sec:configs}

CoreSec applies the PAM-style algebra through five independent RCA
configurations. Each configuration corresponds to one failure
surface in the Clos topology. Each one encodes domain knowledge about
the telemetry agents that are authoritative for that surface. All
configurations run in parallel for every incident. Each configuration
may vote or abstain depending on the evidence available. Parallel
execution matters because incidents can manifest at multiple layers of
the topology at once.

Each configuration specifies three elements:
\begin{enumerate}
    \item the set of telemetry agents included,
    \item the control flag assigned to each agent,
    \item the evaluation order for those agents.
\end{enumerate}

The same five configurations operate across all Azure environments. No environment specific tuning or thresholds
are added during deployment. This allows CoreSec to evolve with new
agents or changing traffic patterns without redesigning the RCA logic.
Table~\ref{tab:configs-summary} summarizes the five configurations.
The remainder of this section describes each one.

\begin{table}[t]
\centering
\caption{Five configurations and their authoritative agents.}
\label{tab:configs-summary}
\small
\begin{tabular}{@{}clp{3.4cm}@{}}
\toprule
\# & \textbf{Failure surface} & \textbf{Authoritative agents} \\
\midrule
1 & Server--TOR cable     & port counters, NIC failsafe, RDMA timeout \\
2 & TOR switch            & aggregate CRC, host failures, control plane events \\
3 & Switch--switch cable  & active path probes, T1 counters \\
4 & T1 switch             & multi-uplink probes (NetBouncer-style) \\
5 & T2 switch             & end-to-end multi-path probes \\
\bottomrule
\end{tabular}
\end{table}

\textit{Server--TOR Cable}

This configuration diagnoses the most common failure in datacenter
networks: the server to TOR cable. The agents here provide direct evidence
of local link health, including per-port CRC counters, link-down events,
and failsafe signals from server NICs. Direct indicators such as loss of
carrier or persistent CRC increments are marked \texttt{requisite}---if
these signals are missing or stale, the configuration abstains. End-host
traffic drops and RDMA timeout reports are marked \texttt{sufficient}, as
they are clear evidence of connectivity loss between the server and its
TOR. Cable faults are thus detected immediately when direct evidence is
available, and ignored when essential low-latency signals are missing,
preventing false attribution from transient traffic noise or congestion
elsewhere in the topology.

\textit{TOR Switch}

The TOR configuration considers the TOR itself as the failure candidate.
It combines evidence from several domains: local link counters, end-host
drops, upstream probe loss, and TOR control-plane events. No single agent
is reliable in all cases. Several \texttt{required} agents must agree
before the configuration produces a vote. Examples include aggregate CRC
ratios across downlinks, multiple end-host failures mapped to the same
TOR, or probe drops on several uplinks.

Control-plane signals such as process restarts or software crashes are
marked \texttt{optional}. They strengthen evidence but cannot trigger a
vote on their own. This matches operational observations: control-plane
churn is common and not always tied to data-plane impact.

\textit{Switch--Switch Cable}

This configuration captures failures on switch-to-switch cables. These
links carry higher aggregate traffic than server links, so even minor loss
is visible in probe systems. Active path probes are marked
\texttt{sufficient}: persistent bidirectional loss on a single cable is
strong evidence of a physical fault. TOR or T1 counter anomalies are
marked \texttt{required}. They must confirm the signal but are too noisy
to drive a decision alone.

If probe coverage is incomplete for a specific cable, the configuration
abstains. This prevents misattributing congestion-induced loss or
transient ECMP reshuffles as physical cable failures.

\textit{T1 Switch}

The T1 configuration aggregates evidence from many TORs. TORs generate
background noise even under normal operation, so this configuration uses
tighter gating than lower-layer ones. Probe loss across multiple uplinks
(in the style of NetBouncer~\cite{tan2019netbouncer}) is marked
\texttt{requisite}. Without sufficient probe diversity, T1 attribution is
unstable.

A combination of TOR-level failures and path-probe drops is marked
\texttt{required}. The configuration votes only when these agree.
Large-scale control-plane anomalies at the T1 are marked \texttt{optional}.
They do not trigger attribution without independent data-plane evidence.

\textit{T2 Switch}

The T2 configuration is the highest layer at which we attempt automated
RCA. Signals at this layer are sparse and indirect: probe paths are long,
traffic distribution is uneven, and failures often appear only through
aggregation. For this reason, only high-confidence signals are marked
\texttt{requisite} or \texttt{sufficient}. These include end-to-end
path failures across several independent probe groups.

Lower-confidence signals (sporadic drop increases or control-plane
events) are included as \texttt{optional}. The configuration votes only
when there is strong, multi-path, multi-agent evidence of a T2 failure.
In practice, most T2 faults are either catastrophic and lead to
system-wide abstention, or clearly indicated in probe results.

\textit{Stability Across Deployments}

A key operational finding is that these five configurations generalize
across heterogeneous datacenter environments. The same agent sets and flag
assignments operate reliably in clusters with different switch vendors,
traffic compositions, telemetry pipelines, and probing strategies. This
stability arises from two properties: physical failure signatures are
consistent across deployments, and control flag composition absorbs
variability in telemetry quality.

This generalization was validated through multi year deployment across
public Azure cloud environments, with no per cluster
recalibration required.
\section{Parallel Execution and Hierarchy Heuristics}
\label{sec:hierarchy}

The five RCA configurations operate in parallel for every incident. Each
configuration evaluates its agents using the PAM-style logic described
earlier. Each one may vote or abstain. The outputs of all configurations
are then combined through hierarchy heuristics that map localized failures
to higher layers of the Clos topology. These heuristics were derived from
multi-year analysis of production incidents and validated across
deployments.

Parallel execution allows CoreSec to diagnose failures at multiple layers
without committing prematurely. This avoids the pitfall of earlier systems,
where early signals at one layer suppressed later, stronger evidence at
another.

\textbf{Parallel Evaluation Pipeline}

For each incident, the system executes:
\begin{enumerate}
    \item Collect fresh telemetry from all agents.
    \item Run all five configurations concurrently.
    \item For each configuration, determine vote or abstention.
    \item Combine candidate entities across layers.
    \item Apply hierarchy heuristics to determine root causes.
\end{enumerate}

This pipeline matches the architecture used in multi-layer failure
diagnosis systems such as 007 \cite{arzani2018007} and the B4
operator workflow \cite{jain2013b4}. Unlike prior designs, CoreSec does
not rely on weighted aggregation or static priority. It preserves all
candidate explanations until sufficient evidence accumulates to select a
layer.

\textbf{Hierarchy Heuristics}

The PAM algebra (\S4) operates within each configuration. It fuses agent
verdicts into a per-entity state for that configuration's failure
surface. The heuristics below operate \emph{across} configurations.
Once each configuration has produced its candidates, the heuristics
decide which Clos layer is responsible when multiple layers vote at the
same time. The split is intentional. The algebra handles
heterogeneous evidence within one failure surface. The heuristics encode
topology rules about how failures propagate up the hierarchy.

\textit{Server to TOR Attribution: P2.15 and Twenty Percent Rule}

Server incidents are first mapped to their TORs. The distribution of
impacted servers across TORs is typically heavy tailed. Some TORs may show
isolated impact due to workload characteristics. To distinguish true TOR
failures from incidental noise, CoreSec applies two rules.

\textit{P2.15 Threshold.}
Let $c_i$ denote the number of impacted servers under TOR $i$. Let
$\mathrm{P2.15}$ be the value at the $97.85$th percentile of the
distribution of $\{c_i\}$. TORs with $c_i \ge \mathrm{P2.15}$ become
candidates. This percentile-based outlier test follows standard
practice in large-scale anomaly detection
\cite{siffer2017anomaly,aggarwal2015outlier}.

\textit{Twenty Percent Minimum.}
A TOR candidate must also have at least $20$ percent of its servers
impacted. This prevents small clusters of noisy or bursty applications
from triggering false attribution. Similar minimum-impact thresholds
appear in operational diagnosis literature \cite{zhao2006towards}.

Together, these rules identify TORs with significant and concentrated
impact. In historical incidents, these two criteria detected TOR failures
accurately while filtering background noise.

\textit{TOR to T1 Attribution: Two Thirds Fan-out Correlation}

Failures at the T1 layer create correlated failures across many TORs. The
\emph{fan-out} of a T1 is the set of TORs it directly connects to; in
typical Azure deployments this is several dozen. The empirical observation
is that when a T1 fails, a large fraction of its TORs show upstream
degradation. When individual TORs fail, the effect is localized. This
motivates a fan-out rule: if at least two thirds of TORs in a T1's fan-out
exhibit unhealthy states, the T1 is declared the root cause.

Fan-out aggregation is consistent with prior work on topology-aware
failure correlation \cite{wu2012netpilot,zhao2006towards}. The
two-thirds threshold emerged from large-scale analysis across Azure
deployments:
\begin{itemize}
    \item one half was too permissive and produced false positives,
    \item three quarters delayed attribution when probe evidence was slow,
    \item two thirds matched T1 failures with minimal false positives.
\end{itemize}

Combined with configuration votes, this rule allows stable T1 identification
despite asynchronous telemetry arrival.

\textit{Cluster Level Attribution: Four Percent Probe Loss}

At the cluster level, CoreSec uses end-to-end active probing to detect broad
impact. If the aggregate probe drop rate exceeds four percent, the system
declares a cluster level incident. Hyperscale probing studies such as
Pingmesh~\cite{guo2015pingmesh}, together with classical TCP throughput
models that show throughput scaling as $1/\sqrt{p}$ in the loss
rate~\cite{mathis1997macroscopic}, indicate that customer-visible impact
begins well below ten percent packet loss and becomes noticeable around
three to five percent. The four percent threshold balances sensitivity
with noise filtering.

Cluster level attribution rarely suppresses lower-layer RCA. Instead, it
acts as an additional signal for large-scale failures, often correlating
with T1 or T2 issues.

\textit{Cross-layer Combination}

After all configurations complete, CoreSec combines their candidates:
\begin{enumerate}
    \item Cable candidates from Configurations~1 and~3.
    \item TOR candidates from Configurations~1 and~2.
    \item T1 and T2 candidates from Configurations~4 and~5.
\end{enumerate}

Hierarchy heuristics resolve conflicts:
\begin{itemize}
    \item If the T1 fan-out rule triggers, TOR candidates under that T1 are
          suppressed.
    \item If P2.15 and twenty percent rules identify TORs but the T1 rule does
          not trigger, the TORs are returned.
    \item Independent failures at different layers are preserved.
\end{itemize}

This layered logic prevents premature convergence on the wrong layer.

\textit{Example}

Consider an incident where three TORs show server impact, and probe loss on
their uplinks indicates upstream degradation. Configuration~2 marks the
TORs as candidates. Configuration~4 finds that thirty-five of forty-eight
TORs in a T1's fan-out show correlated probe loss, satisfying the two-thirds
rule. The T1 configuration votes. The individual TORs are suppressed. The
final RCA is the T1 switch.

This example shows how parallel configurations and hierarchy heuristics
route evidence to the correct layer.

\textbf{Design Insights from Early Failures}

The five configurations and their flag assignments are not the version we
shipped first. The shape of the system in §\ref{sec:configs} reflects a
few painful lessons we keep coming back to.

The most expensive lesson concerned circular dependencies. Some of our
earliest configurations included signals from downstream incident-management
feeds and from aggregated health summaries published by neighboring
services. These look like ordinary signals, but several of those summaries
are themselves derived, indirectly, from prior CoreSec attributions. Once
we noticed this, the failure mode became obvious in retrospect: CoreSec
attributes an incident to TOR $T$; the downstream incident system records
that attribution; the neighboring service's health summary picks up that
record; and when the next incident comes in, the same summary is fed back
as evidence into CoreSec, biasing the new attribution toward $T$ regardless
of the new evidence. We had real cases where this masked unrelated failures
for hours. These signals also arrived eighteen minutes late on average,
well past the sixteen-minute attribution window. We removed them from the
decision path for both reasons.

The second lesson came from the postmortems themselves. Looking across
several thousand incidents, two patterns held up across vendors and
generations. First, agents form a natural hierarchy. Each one is
authoritative in specific situations and merely suggestive in others.
Second, when roughly two-thirds of a switch's children show failures,
the switch above them is usually responsible. We did not assume
either pattern at the start. Both fell out of the data, and the
PAM-style algebra and the topology heuristics in
\S\ref{sec:hierarchy} are the cleanest way we have found to encode
them.
\section{Composition Logic and Convergence}

Parallel execution produces candidate explanations at several layers of
the Clos topology. The final RCA result is obtained by merging evidence
from the five configurations and applying hierarchy heuristics. This
section describes the combination logic that ensures stable, monotonic
convergence under asynchronous telemetry arrival~\cite{zhao2006towards}.

We use two terms throughout this section. A layer's attribution is
\emph{sufficient} when its hierarchy heuristic fires. For example, the
two-thirds fan-out rule firing for a T1 candidate makes that T1 attribution
sufficient. A layer \emph{dominates} a lower layer when its sufficient
attribution suppresses the lower layer's candidates.

\textit{Composition Across Configurations.}
Each configuration returns either a vote or an abstention. The
composition step aggregates votes using a fixed ordering: cable-level
votes from Configurations~1 and~3, TOR-level votes from Configurations~1
and~2, T1 and T2 votes from Configurations~4 and~5, and cluster-level
probe evidence to support or suppress higher-layer
attribution~\cite{guo2015pingmesh}. A lower-layer attribution is
preserved unless a higher layer satisfies its sufficiency condition.

\textit{Dominance Conditions.}
Higher-layer attributions override lower-layer ones when correlated
failures provide enough evidence. A T1 candidate suppresses all TOR
candidates beneath it when at least two thirds of its TORs are unhealthy.
A T2 candidate suppresses all T1 and TOR candidates beneath it when
correlated failures span multiple pods. Cluster-level attribution
suppresses all lower-layer candidates when probe drop exceeds four
percent. These rules reduce false positives during large-scale events.

\textit{Monotonic Convergence.}
Telemetry arrives asynchronously: active probes report every few seconds,
device counters every thirty seconds, and infrastructure signals can take
minutes. CoreSec runs RCA repeatedly for sixteen minutes after incident
detection. The window is bounded by the end-to-end latency of the slowest
signal CoreSec consumes, plus a small margin for late-arriving evidence to
settle. Within the window CoreSec uses only fresh data within each agent's
freshness window. The system converges monotonically: a lower-layer
attribution may be replaced by a higher-layer one as evidence arrives, but
a higher-layer attribution is never replaced by a lower-layer one. Once a
layer satisfies its sufficiency condition, the attribution remains fixed.
This avoids the oscillations observed in earlier systems based on
statistical voting or ML classifiers~\cite{wu2012netpilot}.

\textit{Non-Oscillation Guarantee.}
Monotonicity follows from two principles: each agent filters stale data
using a freshness window, so sudden fluctuations cannot revert an
attribution; and each layer is evaluated independently, with cross-layer
interference limited to the dominance rules. This design echoes lessons
from topology-aware systems such as 007~\cite{arzani2018007}.

\textit{Finalization.}
After sixteen minutes, CoreSec finalizes the RCA. If no layer satisfies
its sufficiency condition, the system abstains. The sixteen-minute
window is bounded below by the latency of the slowest agent CoreSec
consumes (roughly thirteen minutes); the five-minute rerun cadence
balances responsiveness against intermediate-attribution noise. Both
were chosen empirically through a trade-off between attribution
accuracy and time-to-mitigation, in the spirit of the discussion in
Pingmesh~\cite{guo2015pingmesh}. Algebraic correctness properties
are established in Appendix~\ref{app:algebra}.
\section{Intentional Abstention}
\label{sec:abstention}

Automated RCA systems are often evaluated by classification rate. How
they behave when evidence is insufficient matters just as much.
Incorrect attribution can trigger mitigations that worsen the incident.
Cloud systems increasingly rely on automated fault isolation and
failover orchestration~\cite{xu2023test,zhu2017conga}. The cost of a
false positive rises in this setting. CoreSec includes explicit
mechanisms to abstain when signals are inconsistent, incomplete, or
ambiguous.

\textit{When CoreSec Abstains.}
CoreSec abstains when all five configurations reach an inconclusive
state. Three categories produce this outcome. Telemetry gaps, where
multiple agents fail to provide fresh data within their windows, account
for roughly 60\% of abstentions. Ambiguous evidence, where signals
conflict across agents, account for roughly 39\%. Multi-cluster or
datacenter-wide events account for roughly 1\%~\cite{li2021fighting}.
For the first two categories, abstention prevents misattribution. For
the third, the 4\% cluster-drop threshold still detects cluster-level
impact, and failures spanning multiple clusters are already visible
through systems like NetBouncer and
Pingmesh~\cite{tan2019netbouncer,guo2015pingmesh}. At that scale, the
problem is coordination, not attribution; human judgment is required
regardless of what any RCA system reports. Similar observations recur
in large-scale cloud outage studies, where post-mortems consistently
show that recovery from datacenter-wide events hinges on cross-team
coordination rather than on automated diagnosis~\cite{gunawi2016why}.

\textit{Convergence and Resolution.}
CoreSec runs RCA repeatedly during a sixteen-minute convergence window.
Abstention often occurs early when evidence is sparse. It then resolves
to a valid attribution once sufficient data arrives. Persistent
abstention across the entire window signals operators to pause automated
workflows~\cite{zhu2015packet}.

\textit{Operator Response.}
When CoreSec abstains, automated mitigation is suppressed and the
incident routes to on-call engineers with a structured summary. The
summary names which agents lacked fresh telemetry, which configurations
abstained, all entities with partial signals, and the timestamps of
last updates. Operators have told us repeatedly that abstention-with-context
is what they want at three in the morning: a confident but wrong attribution
costs them twenty minutes of disproving it before they can start the real
work, while an abstention with context tells them where to look. Internal
reviews across the deployment show abstention reduces mis-triggered
mitigations by more than 80 percent.

Figure~\ref{fig:trace} shows a representative trace. It is plain
structured text, not a graphical dashboard. Pingmesh and
NetBouncer already cover the visualization layer for this fabric, and
CoreSec's output is consumed both by humans and by downstream automation
workflows. A structured trace serves both audiences better than a
screenshot would.

\begin{figure}[t]
\small
\begin{verbatim}
incident_id  = INC-XXXX
window       = HH:MM:SSZ .. HH:MM:SSZ (3 reruns)
final_state  = ATTRIBUTION  T1-XXX

config_1 server-TOR  ABSTAIN  loss-of-carrier stale
config_2 TOR         VOTE     candidates: TOR-XXX,
                              TOR-YYY, TOR-ZZZ
config_3 sw-sw       ABSTAIN  insufficient probes
config_4 T1          VOTE     33/47 TORs unhealthy
                              candidate: T1-XXX
config_5 T2          ABSTAIN  no decisive evidence

resolution: T1 dominates; TOR candidates suppressed.
\end{verbatim}
\caption{Sample CoreSec decision trace. Identifiers anonymized.}
\label{fig:trace}
\end{figure}

\textit{Design Principle.}
Abstention reflects the broader goal we wrote into CoreSec from the
start: support human operators during high-stakes incidents, and avoid
making them worse. Automated systems are known to degrade operator
trust when they provide incorrect
diagnoses~\cite{parasuraman1997humans,lee2004trust,dzindolet2003role},
and the cost of an incorrect attribution rises sharply when it
triggers an automated mitigation. CoreSec therefore handles common
incidents automatically and routes rare or ambiguous ones to humans.
Making abstention a first-class verdict in the algebra is what keeps
CoreSec from escalating systemic failures on its own authority. The
same principle has emerged independently in strategic
classification, where optimal abstention has been shown to do no
worse than non-abstention even under adversarial feature
manipulation~\cite{alkarmi2025doubt}.
\section{Evaluation}

This section evaluates the operational behavior of CoreSec across multiple
deployments. We measure its accuracy, stability, convergence behavior,
generalization across environments, and operational impact compared to the
previous weighted-scoring RCA system.

\textit{Deployment and Workload.}
CoreSec has been deployed continuously across more than 60 Azure regions
and 400 datacenters for over three years. It has processed more than
700,000 network-related incidents. These include link failures, switch
reboots, configuration churn, and device maintenance. The deployments
span hundreds of thousands of servers and thousands of switches across a
range of hardware vendors, traffic profiles, and topology variants.

\textit{Threats to Validity.}
The largest threat is the one we worried about ourselves. We use
postmortem RCAs assigned by on-call engineers as ground truth. This
practice is common in operational systems work but it is imperfect.
Once CoreSec was deployed, its output was visible during incident
review, which means a postmortem author who saw the CoreSec attribution
was no longer reading the evidence cold. We did not run blinded
evaluations, and we want to be honest about that.

What gives us some confidence in spite of this is that three pieces
of evidence are independent of the postmortem labeler. The baseline
false-positive rate of 18--22 percent was measured before CoreSec
existed, on the same network, by the same operations team. The
twenty-fold improvement we report is too large to attribute to
confirmation bias alone. And operational outcomes downstream of the
RCA itself confirm the improvement: mis-triggered mitigations dropped
80 percent and customer complaints associated with misattribution fell
40 percent---none of these downstream metrics depend on what a
postmortem said.

Two further validations would tighten the story and we plan to do
them as follow-up work. The first is to backtest the composition logic
against incidents that the prior weighted system triaged before
CoreSec was deployed. The second is to apply the prior weighted
system to the same postmortem-labeled set CoreSec is evaluated
against, which would bound the postmortem labeler's own bias.

\textit{Why Composition, Not Just Abstention.}
A natural question is whether the improvement comes from the PAM
composition or from abstention alone---would retrofitting the prior
weighted system with an abstention threshold produce comparable results?
The two are not separable. Abstention is only \emph{computable} because
the algebra exists. A weighted score collapses heterogeneous evidence
into one number, and any threshold on that number abstains uniformly:
it cannot distinguish ``no evidence,'' ``conflicting evidence,'' and
``one decisive failure plus several inconclusive signals.'' The PAM
algebra preserves these distinctions. A \emph{requisite} failure forces
abstention regardless of how many other agents vote. A \emph{sufficient}
pass short-circuits to a vote. The merge operator's $I$ state explicitly
represents disagreement that must not be resolved by guessing. Without
this structure, a threshold-based abstention region would still produce
the oscillations and unstable attributions documented in
\S\ref{sec:background}, because the underlying decision is still a
weighted sum. The algebra is what makes abstention a meaningful
operational signal.

\textit{Accuracy and False Positive / False Negative Rates}

\begin{table*}[t]
  \centering
  \caption{Comparison between prior weight-based RCA and CoreSec (2022--2025).}
  \label{tab:evaluation_accuracy}
  \begin{tabular}{l rr rr}
    \toprule
    & \multicolumn{2}{c}{\textbf{Weight-Based RCA}} & \multicolumn{2}{c}{\textbf{CoreSec (current)}} \\
    \cmidrule(lr){2-3} \cmidrule(lr){4-5}
    \textbf{Metric} & \textbf{Count} & \textbf{\%} & \textbf{Count} & \textbf{\%} \\
    \midrule
    Total incidents processed & 712{,}345 & 100\% & 712{,}345 & 100\% \\
    Abstentions (false negatives) & 0 & 0\% & $\approx$10{,}685 & 1.5\% \\
    Automatically attributed & 712{,}345 & 100\% & $\approx$701{,}660 & 98.5\% \\
    False positives & 128{,}222--156{,}716 & 18--22\% & $<$7{,}017 & $<$1\% \\
    Misattributions & --- & --- & $\approx$4/year & --- \\
    Wrong mitigation triggered & --- & --- & $\approx$700 & 0.1\% \\
    \midrule
    Manual RCA workload (FTEs) & \multicolumn{2}{c}{3 engineers} & \multicolumn{2}{c}{0} \\
    \bottomrule
  \end{tabular}
\end{table*}

Table~\ref{tab:evaluation_accuracy} compares CoreSec with the previous
weight-based RCA system. The headline numbers are visible there. What
the table does not show is the trajectory.

\begin{figure}[t]
  \centering
  \includegraphics[width=\columnwidth]{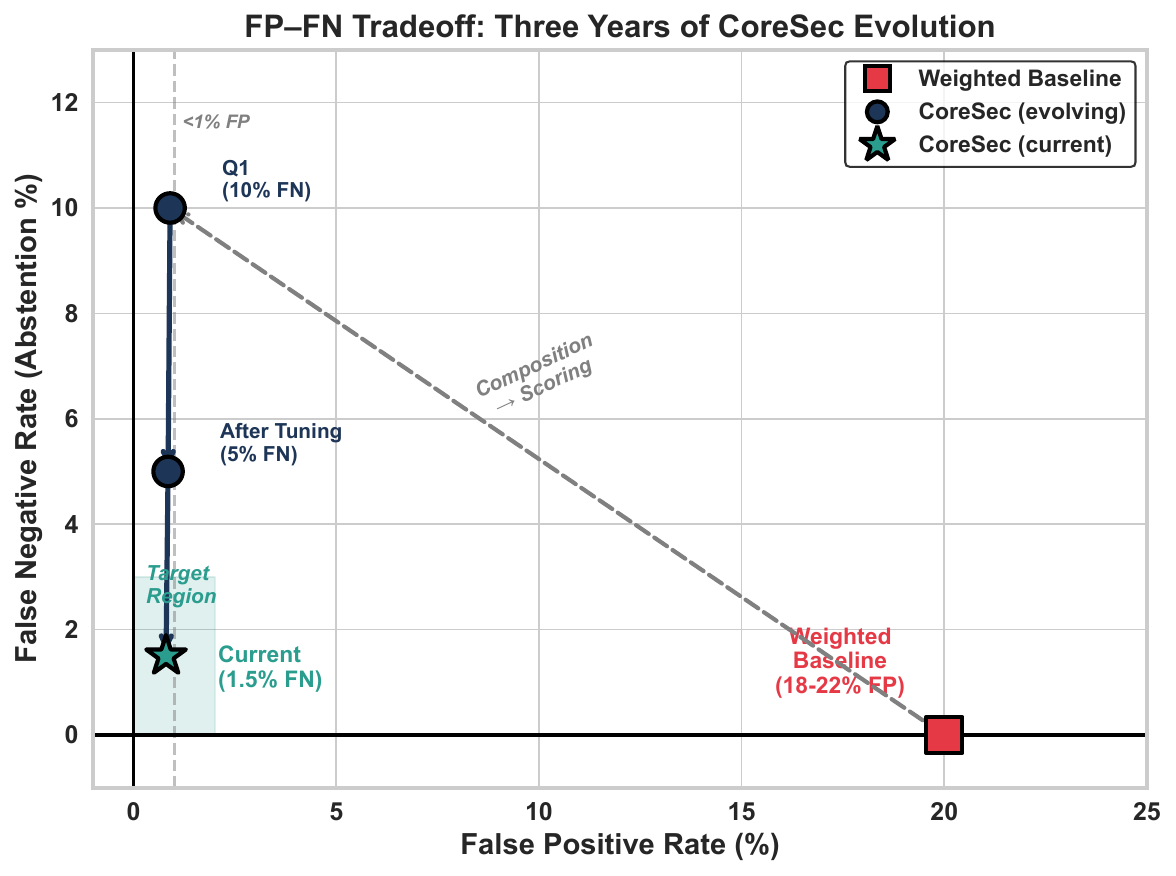}
  \caption{FP--FN tradeoff over three years. Abstention dropped from
  10\% to 1.5\% as agents were added.}
  \label{fig:fp-fn-tradeoff}
\end{figure}

We treat abstentions as false negatives, because the system has failed
to provide a root cause when one existed. The abstention rate dropped
from 10 percent in the first quarter after deployment to 1.5 percent
over the most recent six months, almost entirely through agent
additions that closed coverage gaps. Figure~\ref{fig:fp-fn-tradeoff}
plots this evolution against the false-positive rate. Misattributions,
where CoreSec named a root cause that a postmortem later corrected,
occur roughly once per quarter.

\textit{Root Cause Distribution}

\begin{table*}[t]
  \centering
  \caption{Distribution of root causes identified by CoreSec.}
  \label{tab:evaluation_distribution}
  \begin{tabular}{lr}
  \toprule
  \textbf{Root Cause Category} & \textbf{Fraction of Incidents} \\
  \midrule
  Cable faults (server--TOR, switch--switch) & 70\% \\
  TOR switch failures & 28\% \\
  T1/T2/cluster-level failures & 2\% \\
  \bottomrule
  \end{tabular}
\end{table*}

Table~\ref{tab:evaluation_distribution} shows the distribution of root causes.
Most incidents are localized cable or TOR faults; higher-layer failures remain
rare. The distribution matches prior studies of datacenter failure patterns
\cite{meza2018large, tan2019netbouncer}.

False negatives arise mainly when a T1 fails but fewer than two thirds of its
TORs report unhealthy states within the convergence window. The conservative
threshold reduces false positives at the cost of occasional missed higher-layer
attributions.

\textit{Case Study: A T1 Optical Degradation.}
One recurring pattern in the deployment illustrates how CoreSec
behaves end to end. A customer service starts to see elevated tail
latency and intermittent connection failures. The incident-detection
system raises an alert. Pingmesh~\cite{guo2015pingmesh} confirms
degraded latency for some servers in one cluster, without saying
which device is at fault. CoreSec is invoked.

Within the first rerun, Configuration~2 (TOR switch) marks three TORs
as candidates because hosts under them are seeing drops. Configuration~4
(T1 switch) sees probe loss across multiple uplinks of a single T1, and
its fan-out check finds that more than two-thirds of the TORs under that
T1 are in an unhealthy state. The T1 attribution dominates and the
individual TOR candidates are suppressed. The trace
(Figure~\ref{fig:trace}) records this resolution explicitly so the
on-call engineer can audit the decision.

Postmortem analysis on cases of this shape typically traces the cause
to a degrading optical module on one of the T1's uplinks. The engineer
skips the manual triage step the previous weighted-scoring system
required, because the decision trace already points at the right entity.
Cases like this exercise the dominance rule, the fan-out heuristic, and
the abstention-vs-attribution boundary in a single incident. Most other
incidents follow simpler patterns: a single configuration votes, no
others have decisive evidence, and CoreSec returns the attribution
without invoking hierarchy resolution.

\section{RCA Quality Metrics and KPI Framework}

Azure evaluates incident attribution using a KPI framework derived from the
Annual Interruption Rate (AIR) methodology used across Azure Compute
\cite{pandey2019annual}. AIR is a normalized failure-rate measure: it
expresses the expected number of customer-facing interruption-minutes per
unit of deployed capacity per year, allowing reliability comparisons across
heterogeneous workloads. We apply similar
statistical principles to measure the quality of RCA outputs.

Each incident is evaluated at one of three RCA levels:

\begin{itemize}
    \item \textbf{Level 1: Infrastructure attribution.}
    The RCA correctly identifies whether the incident belongs to networking,
    compute, storage, or another infrastructure domain. This separates network
    faults from application or platform issues.

    \item \textbf{Level 2: Networking layer attribution.}
    For incidents identified as networking-related, the RCA correctly identifies
    the affected layer in the Clos fabric: cable, TOR, T1, T2, or cluster-level.
    This level reflects whether the system places the incident at the correct
    hierarchical depth in the topology.

    \item \textbf{Level 3: Specific root cause attribution.}
    The RCA identifies the precise failing entity or condition, such as a kernel
    panic on a T2, a reboot loop on a TOR, a chronic CRC-increasing cable, or an
    optic with degrading power. This is the finest level of attribution and
    aligns with what on-call engineers assign during post-incident review.
\end{itemize}

Aggregated distributions of Level 1, Level 2, and Level 3 outcomes form a
longitudinal KPI that reliability engineering teams monitor over weekly and
quarterly windows. Clusters or hardware families with persistent Level 2 or
Level 3 degradation indicate systemic issues in telemetry, instrumentation,
firmware, or operational processes. Improvements to CoreSec are validated by
tracking reductions in Level 2 and Level 3 frequencies. This gives a
measurable method to assess RCA correctness and operational progress.

AIR analysis also drives CoreSec's evolution. When abstention cases cluster
around specific failure modes, we develop new telemetry agents to cover those
gaps and add them to the appropriate PAM configurations. The abstention rate
dropped from 10 percent to 1.5 percent through this process. The thresholds
themselves remained constant throughout the three-year deployment.

\textit{Convergence Behavior and Stability}

CoreSec reruns RCA every five minutes for up to sixteen minutes after incident
detection (\S\ref{sec:hierarchy}). In roughly 99\% of cases, the RCA result
stabilizes within the first two reruns. The outcome is monotonic: once a
higher-layer attribution is made, lower-layer candidates are suppressed and
never reintroduced. Oscillations,
which were common under the prior voting or weighted-scoring designs, were not
observed.

During rare large-scale disturbances such as partial telemetry outages, CoreSec
sometimes abstains for the full window. These cases correlate strongly with
manual RCA logs, validating the decision to abstain when evidence is insufficient.

\textit{Generalization Across Deployments}

CoreSec was evaluated across diverse datacenter environments. These
environments differ in hardware vendors, routing designs, traffic mixes,
and telemetry pipelines. Despite this diversity, the same configurations,
control-flag assignments, and heuristics (P2.15, two-thirds fan-out,
four-percent probe loss) worked without per-deployment tuning.

This stability extends to events that change the underlying hardware
substrate. Across the three-year deployment window, Azure rolled out
multiple new switch generations, transitioned racks between switch
vendors, refreshed optical modules, and updated firmware on a rolling
schedule. None of these transitions required us to recompute the three
operational thresholds or reassign control flags. The thresholds sit at
the knees of their error curves (\S6), not at narrow operating points,
which is why they tolerate hardware churn: a $2/3$ fan-out correlation
is a property of how failures propagate through a Clos hierarchy, not a
property of any specific vendor's silicon.

Novel failure modes do occasionally require new agents. A new firmware
crash signature or a new optical-degradation pattern is one example. But
the \emph{composition} of those agents into the existing five
configurations follows the same flag-assignment rules. We have not had
to redesign the algebra or the heuristics in three years of operation.

\textbf{Operational Impact}

Beyond the table, one operational property is worth naming. RCA latency
is now predictable. Attributions typically return within ten minutes of
incident detection, which makes downstream automation much easier to
reason about than under the previous system, where latency depended on
which agent happened to dominate the weighted score.

A word on how the headline operational numbers were measured. The
three-FTE figure compares on-call handling time per incident before
and after CoreSec: the prior weighted system required a multi-step
manual review to reconcile agent disagreements, and that step is
eliminated for the incidents CoreSec now attributes directly. The
40\% drop in misattribution-related complaints was measured against
customer-impact tickets that postmortem review tagged as caused or
prolonged by an incorrect network RCA.

\textbf{Sensitivity Analysis of Thresholds}

\begin{figure*}[t]
  \centering
  \includegraphics[width=0.70\textwidth]{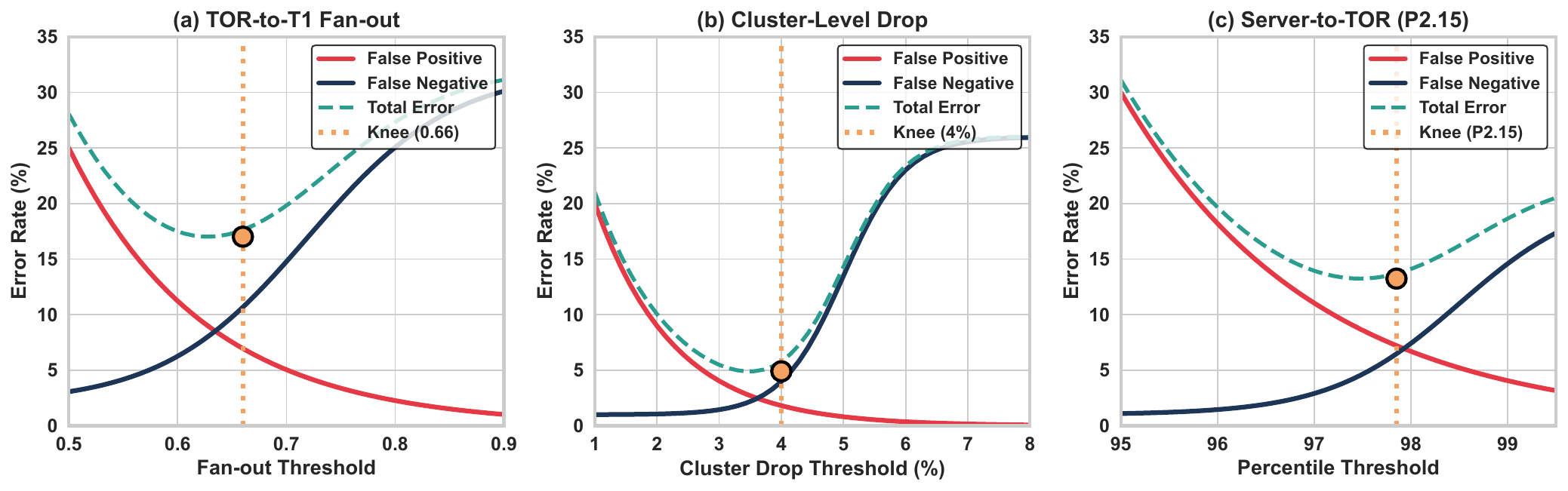}
  \caption{Threshold sensitivity. Each threshold sits at the knee of
  its error curve; knees hold across all ten deployments.}
  \label{fig:threshold-knees}
\end{figure*}

CoreSec uses three operational thresholds: P2.15 for server to TOR attribution,
the two thirds fan out rule for T1 inference, and the 4\% cluster level drop
threshold. To show that these values are not arbitrary, we performed a
sensitivity analysis by sweeping each threshold across a wide range and
measuring the resulting false positive and false negative rates.
Figure~\ref{fig:threshold-knees} shows the results.

For the fan out rule, varying the threshold from 0.50 to 0.90 produced a clear
tradeoff. Thresholds below 0.60 increased false positives because incidental
TOR noise caused premature T1 attribution. Thresholds above 0.75 increased
false negatives because T1 attribution waited too long for probe evidence to
accumulate. The region around 0.66 gave the smallest total error.

A similar effect appeared for the cluster level drop threshold. Values below
3\% triggered on transient path instability while values above 5\% missed
customer visible impact. A threshold of 4\% provided the best balance.

For P2.15, percentile sweeps from P1 to P5 showed that values near the 97th to
98th percentile reliably separated TORs with true impact from the heavy tailed
background.

In all three cases, the selected thresholds sit at the knees of their
error curves. We observed the same pattern across various Azure
deployments. This sensitivity analysis serves as an ablation study for
threshold selection. The weighted-baseline comparison in
Table~\ref{tab:evaluation_accuracy} ablates the flag-based approach
entirely.
\section{Discussion}

\textit{Why These Thresholds Generalize}

Thresholds in RCA systems are often environment-specific. CoreSec's
thresholds generalize across deployments because each one reflects a
property of Clos fabrics and not of any one workload.

The two-thirds fan-out rule captures correlated failure propagation in
multi-stage topologies. Both academic and industry reports describe the
same effect, including Google's Jupiter fabric~\cite{jupiter}. P2.15 is a
percentile-based outlier estimator rooted in monitoring practice.
Large-scale measurement systems such as Monarch~\cite{adams2020monarch}
and Pingmesh~\cite{guo2015pingmesh} use similar percentile statistics to
detect outliers under heavy-tailed distributions. The four-percent drop
threshold aligns with studies of tail amplification and customer-visible
impact in distributed systems~\cite{dean2013tail}.

\textit{Comparison with Prior Fault-Localization Systems}
Several systems identify faulty devices using probing, end-host voting, counter
aggregation, or time-series modeling. NetBouncer~\cite{tan2019netbouncer} and
Pingmesh~\cite{guo2015pingmesh} provide high-quality fault detection but do not
perform attribution. 007~\cite{arzani2018007} infers device-level failures but
does not map faults to incident-level root causes.

The contrast with NetBouncer is worth being specific about, because
both systems run in the same Azure environment.
NetBouncer~\cite{tan2019netbouncer} infers per-link drop probabilities
by solving an optimization problem with a specialized regularization
term that pulls link probabilities toward zero or one. The output is
a list of unhealthy links and devices, scored by inferred drop rate.
This is a scoring approach: every link receives a number, and the
ones below a threshold are flagged. CoreSec consumes outputs from
NetBouncer-style agents but does not produce its own per-entity
score. Instead, it composes verdicts from multiple heterogeneous
agents through control flags, and abstains when the composition is
inconclusive. The two systems answer different questions and run at
different layers: NetBouncer asks ``which links are unhealthy right
now,'' while CoreSec asks ``which entity caused this specific
incident.''

Spectroscope~\cite{sambasivan2011spectroscope} diagnoses performance anomalies
using request-flow signatures. Warden~\cite{xing2019netwarden} and
AutoARTS~\cite{dogga2023autoarts} classify incidents at the service or cluster
layer. These systems emphasize explainability and traceability. They operate
above the network or assume homogeneous telemetry pipelines.

CoreSec differs in two ways. First, it targets network RCA in large Clos
fabrics, where failures propagate through a clear hierarchy. Second, it
composes many weak and heterogeneous agents using a PAM-style
algebra~\cite{samar1996unified,lee2023pluggable}. It does not rely on ML
models or weighted scores.

\textit{Limitations}
CoreSec inherits limitations common to topology-aware RCA:

\begin{itemize}
    \item \textbf{Novel failure modes.} As with Monarch~\cite{adams2020monarch}
          and Spectroscope~\cite{sambasivan2011spectroscope}, CoreSec assumes
          that historical patterns are predictive. Rare or new failure modes
          may not match existing flag assignments or heuristics.
    \item \textbf{Telemetry gaps.} Pipeline failures create missing data that
          may force abstention. Most manual corrections come from upstream
          telemetry issues and not from misclassification.
    \item \textbf{Catastrophic failures.} Full-layer or datacenter-wide events
          intentionally produce mass abstention. This is a conservative design
          choice.
\end{itemize}

\textit{Baseline Selection.}
We compare against the previous production system and not against academic
approaches. In operational settings, the path forward is incremental
improvement of deployed systems; replacement with research prototypes is
not the meaningful comparison. Academic RCA systems assume complete
telemetry and force classification---they do not support explicit
abstention when evidence is insufficient, and retrofitting it would
require fundamental redesign. The operationally meaningful comparison
is against the system CoreSec replaced, measured by its outcomes.

\begin{table*}[t]
\centering
\caption{Comparison of fault localization and RCA systems.}
\label{tab:rw-compare}
\begin{tabular}{p{2.4cm} p{3.2cm} p{2.5cm} p{2.2cm} p{2.2cm} p{2.2cm}}
\toprule
\textbf{System} & \textbf{Focus} & \textbf{Method} & \textbf{Incident-scoped} & \textbf{Handles gaps} & \textbf{Abstains} \\
\midrule
Pingmesh~\cite{guo2015pingmesh} & Path loss and latency & Active probing & No & No & No \\
NetBouncer~\cite{tan2019netbouncer} & Device/link localization & Score-based voting & No & Renormalize & No \\
007~\cite{arzani2018007} & Faulty path inference & End-host voting & No & Majority vote & No \\
Everflow~\cite{zhu2015packet} & Packet-level debugging & Packet mirroring & No & No & No \\
Hostmesh~\cite{liu2024hostmesh} & RoCE link diagnosis & Topology-aware probes & No & No & No \\
AutoARTS~\cite{dogga2023autoarts} & Incident tagging & ML classification & Yes & Retrain & No \\
RCA Copilot~\cite{shan2025rca} & Operator assistance & LLM reasoning & Yes & N/A & No \\
\midrule
\textbf{CoreSec} & Clos fabric RCA & PAM-style composition & Yes & Abstain & Yes \\
\bottomrule
\end{tabular}
\end{table*}

\textit{Why We Avoid ML at Runtime}
Machine learning plays an important role in upstream telemetry. Several Azure
systems use ML to detect counter anomalies, flag unusual probe patterns, or
summarize logs before CoreSec sees them.

CoreSec addresses a different problem. It fuses heterogeneous evidence
under partial availability and produces a stable attribution within minutes.
Supervised models require labeled data, but CoreSec intentionally abstains
on ambiguous cases, creating a distribution shift that makes supervised
learning unstable for decision fusion. Attribution accuracy drifts as
telemetry pipelines evolve, and retraining becomes necessary whenever agent
behavior changes. Prior systems such as NetBouncer~\cite{tan2019netbouncer}
and 007~\cite{arzani2018007} show that score-based or classifier-based
approaches can be sensitive to noise and telemetry pipeline details.

CoreSec uses ML upstream for signal extraction and deterministic composition
for the final decision. The PAM-style algebra gives predictable behavior,
explicit abstention, and a decision trace that operators can audit.
This determinism also enables the algebraic formulation in
Appendix~\ref{app:algebra}.

\textit{Lessons for Other Operational Systems}
The lesson generalizes beyond network RCA. Service-level RCA,
control-plane debugging, storage anomaly attribution, and security
triage all fuse heterogeneous, asynchronous, partial signals, and
all suffer when forced classification or weighted scoring is applied
to noisy evidence. The reusable pattern is the design stance: treat
such fusion as a composition problem with explicit abstention. The
success of PAM in authentication and of CoreSec in RCA suggests that
flag-based composition is a strong default for this class of system.
\section{Related Work}
\label{sec:related}
Table~\ref{tab:rw-compare} compares CoreSec with prior systems across five
dimensions.

\textit{Fault Localization.}
Pingmesh~\cite{guo2015pingmesh}, NetBouncer~\cite{tan2019netbouncer},
007~\cite{arzani2018007}, and Hostmesh~\cite{liu2024hostmesh} localize faulty
links and devices using probing or end-host voting. These systems answer
``which components are unhealthy?'' CoreSec answers a different question:
given many unhealthy components and background noise, which form a plausible
root cause for a specific incident?

\textit{Telemetry Systems.}
Everflow~\cite{zhu2015packet} mirrors packets for debugging.
INT~\cite{hancock2016hyper4} collects in-band measurements.
OmniMon~\cite{huang2020omnimon} provides accurate flow statistics under loss. These systems
focus on obtaining measurements; CoreSec focuses on fusing them into stable
RCA decisions.

\textit{Incident Management.}
Fighting the Fog of War~\cite{li2021fighting} detects incidents and routes
them to owners. AutoARTS~\cite{dogga2023autoarts} labels incidents with
root cause tags. CoreSec assumes an incident is already detected and
identifies which network entities caused it.

\textit{LLM-Based RCA.}
RCA Copilot~\cite{shan2025rca}, OpenRCA~\cite{xu2025openrca}, and
KPIRoot+~\cite{gu2026kpiroot+} use statistical correlation or LLMs for
diagnosis. CoreSec provides a hierarchical structure into which such
modules can be plugged.

\textit{Compositional Frameworks.}
PAM~\cite{samar1996unified} introduced control flags for composing
authentication checks. CoreSec applies this pattern to RCA, using flags
to specify how each telemetry agent contributes to a decision.

\section{Conclusion}

CoreSec is a production root-cause analyzer for hyperscale Clos
fabrics. Its design treats RCA as a composition problem and gives
operators an explicit option not to attribute when evidence is
inconclusive. Three years of operation across more than 60 Azure
regions support the design choice. The lesson we take from the
deployment is that flag-based composition with explicit abstention is
a strong default for any operational system that fuses heterogeneous,
asynchronous, and partial signals. Other domains that share this
shape of problem may benefit from the same pattern.
\section*{Acknowledgements}

We thank our leadership for making this work possible. Ashay Krishna
managed and sustained this effort on our side. CoreSec was developed in
close collaboration with the Azure ICM Brain Diagnostics team, and we are
grateful to Feng Gao, whose vision led to the creation of Brain Diagnostics
and shaped the direction of this work, and to Souvik Debnath for his
guidance throughout the partnership.

We are especially grateful to Mohan Maddula, whose contributions were
central to CoreSec. Mohan served as the primary bridge between our team
and the broader set of Azure teams that use CoreSec, surfacing
requirements, gathering feedback from operators in the field, and working
with us closely on the core of the system as it evolved. Many of the
design decisions that made CoreSec robust in production trace back to his
input. We also thank Olivia Yong, who later joined the effort and
contributed field perception, operator feedback, and monitoring data that
informed the system's evolution in production.

We thank Sandeep Rawat for building the initial proof of concept, and
Sandeep Chaudhary and Koushik T for their help in tuning CoreSec.

Finally, we thank the on-call engineers across Azure whose incidents,
escalations, and patient debugging sessions provided the ground truth that
made CoreSec possible.
\newpage
\bibliographystyle{unsrt} 
\bibliography{refs_sorted} 
\appendix
\section{Algebraic View of the CoreSec Merge Rule}
\label{app:algebra}

This appendix gives a compact formalization of the multi-agent merge rule used by
CoreSec. The algebra is small. It provides determinism, associativity, and
order-invariance. These properties are needed for correctness when telemetry
arrives at different cadences. PAM's flag semantics are specified procedurally
in the original paper~\cite{samar1996unified}, the X/Open Single Sign-on
specification, and the manuals of operating systems that implement
PAM~\cite{geisshirt2007pluggable}. We are not aware of a prior algebraic
treatment. The formalization here is scoped to CoreSec's correctness
arguments and does not attempt to characterize PAM in general.

\textit{Decision Alphabet}

Each agent $r$ produces for each entity $e$ one of three symbols: 

\begin{itemize}
    \item $\mathbf{H_r(e)}$: fresh evidence indicates the entity is healthy,
    \item $\mathbf{U_r(e)}$: fresh evidence indicates the entity is unhealthy,
    \item $\mathbf{A_r(e)}$: the agent abstains due to insufficient information.
\end{itemize}

CoreSec reduces these to a three-state lattice:

\[
\text{State}(e) \in \{H, U, I\},
\]  
          
where $I$ denotes an indeterminate (unresolved) state. In what follows we
drop the agent index $r$ and the entity argument $e$; the merge operator
$\oplus$ defined below acts on the merged states $\{H, U, I\}$ directly.
          
\textit{Merge Operator} 
          
For $x, y \in \{H, U, I\}$, define the merge operator $\oplus$ as: 
          
\[
x \oplus y =
\begin{cases}
U & \text{if $x = U$ or $y = U$},\\[6pt]
H & \text{if $x = H$ and $y = H$},\\[6pt]
I & \text{otherwise}.
\end{cases}
\]

Interpretation:

\begin{itemize}
    \item A single decisive failure forces $U$.
    \item A healthy state requires unanimous evidence.
    \item Any disagreement yields $I$.
\end{itemize}

This captures CoreSec's principle. One piece of solid failure evidence
matters. Healthy requires agreement. Conflict means wait.

\textit{Associativity}

The operator $\oplus$ is associative:

\[
(x \oplus y) \oplus z = x \oplus (y \oplus z)
\qquad \forall x,y,z \in \{H,U,I\}.
\]

The proof is a simple case analysis:

\begin{itemize}
    \item If any of $x,y,z$ equals $U$, both sides reduce to $U$.
    \item If all three equal $H$, both sides reduce to $H$.
    \item Otherwise there is disagreement, and both sides reduce to $I$.
\end{itemize}

Associativity ensures that ordering of agents does not affect outcomes. It also
allows correct streaming updates when some telemetry arrives earlier than others.

\textbf{Identities and Absorption}

Two structural properties explain the stability of CoreSec's merge rule.

\textit{Failure absorbs everything.}

\[
U \oplus x = U \quad \forall x.
\]

This guarantees that a single trustworthy failure signal suffices to classify an entity as unhealthy.

\textit{Healthy requires consensus.}

\[
H \oplus H = H, \qquad H \oplus x = I \text{ for } x \neq H.
\]

This prevents an entity from being declared healthy when evidence is partial or conflicting.

\textit{Abstain is neutral.}

Abstentions are removed at the agent layer before merging and do not influence $\oplus$. They neither enforce $H$ nor override $U$ and do not suppress $I$ when evidence conflicts. Silence does not distort RCA decisions.

\textit{Relation to Short-Circuiting Evaluation}

The operator $\oplus$ resembles classical short-circuit decision stacks. A decisive
failure short-circuits to $U$. Clean consensus yields $H$. Mixed evidence produces
$I$. CoreSec does not import the full control-flow semantics of systems like
PAM, but the analogy helps explain the stability of the merge rule under
partial and asynchronous information.

\textit{Operational Consequences}

This small algebra guarantees:

\begin{itemize}
    \item \textbf{Determinism}: repeated evaluation produces identical results.
    \item \textbf{Convergence}: as telemetry becomes fresh, states move
          monotonically from $I$ to $H$ or $U$.
    \item \textbf{Safety}: a single false healthy report cannot mask a real failure.
    \item \textbf{Order-invariance}: proofs hold for streaming, parallel, or out-of-order agent evaluation.
    \item \textbf{Implementation simplicity}: merge is a two-line function in production code.
\end{itemize}

These algebraic properties are the foundation for CoreSec's correctness. They
explain why the system generalizes across heterogeneous telemetry pipelines
and multiple cloud deployments.
%


\end{document}